\documentclass[11pt]{article}
\usepackage[margin=1in]{geometry}
\usepackage{booktabs}
\usepackage{amsmath}
\usepackage{authblk}
\usepackage{hyperref}
\newcommand{\ci}[2]{[\text{\tiny #1,#2}]}
\renewcommand{\thefootnote}{\arabic{footnote}}

\usepackage[T1]{fontenc}
\usepackage{graphicx,verbatim}
\begin{document}
\title{CRIMSON: A Clinically-Grounded LLM-Based Metric for Generative Radiology Report Evaluation}

\date{}

%

\author[1,*,$\dagger$]{Mohammed Baharoon}\author[2,*]{Thibault Heintz}
\author[1,*]{Siavash Raissi}
\author[3]{Mahmoud Alabbad}
\author[4]{Mona Alhammad}
\author[5]{Hassan AlOmaish}
\author[1]{Sung Eun Kim}
\author[1]{Oishi Banerjee}
\author[1]{Pranav Rajpurkar}

\affil[1]{Department of Biomedical Informatics, Harvard Medical School, Boston, MA}
\affil[2]{Department of Radiation Oncology, Mass General Brigham, Boston, MA}
\affil[3]{King Fahad Hospital, Al-Ahsa Health Cluster, Al Hofuf, Saudi Arabia}
\affil[4]{Ras-Tanura General Hospital, Ministry of Health, Eastern Province, Saudi Arabia}
\affil[5]{Department of Medical Imaging, King Abdulaziz Medical City, Ministry of National Guard, Riyadh, Saudi Arabia}
\affil[*]{These authors contributed equally}

\maketitle              

\begingroup
\renewcommand{\thefootnote}{$\dagger$}
\footnotetext{Contact: MohammedSalimAB@outlook.com}
\endgroup

\begin{abstract}
We introduce \textbf{CRIMSON}, a clinically grounded evaluation framework for chest X-ray report generation that assesses reports based on diagnostic correctness, contextual relevance, and patient safety. Unlike prior metrics, CRIMSON incorporates full clinical context, including patient age, indication, and guideline-based decision rules, and prevents normal or clinically insignificant findings from exerting disproportionate influence on the overall score. The framework categorizes errors into a comprehensive taxonomy covering false findings, missing findings, and eight attribute-level errors (e.g., location, severity, measurement, and diagnostic overinterpretation). Each finding is assigned a clinical significance level (urgent, actionable non-urgent, non-actionable, or expected/benign), based on a guideline developed in collaboration with attending cardiothoracic radiologists, enabling severity-aware weighting that prioritizes clinically consequential mistakes over benign discrepancies. CRIMSON is validated through strong alignment with clinically significant error counts annotated by six board-certified radiologists in ReXVal (Kendall’s $\tau = 0.61$–$0.71$; Pearson’s $r = 0.71$–$0.84$), and through two additional benchmarks that we introduce. In \textit{RadJudge}, a targeted suite of clinically challenging pass–fail scenarios, CRIMSON shows consistent agreement with expert judgment. In \textit{RadPref}, a larger radiologist preference benchmark of over 100 pairwise cases with structured error categorization, severity modeling, and 1–5 overall quality ratings from three cardiothoracic radiologists, CRIMSON achieves the strongest alignment with radiologist preferences. We release the metric, the evaluation benchmarks, RadJudge and RadPref, and a fine-tuned MedGemma model to enable reproducible evaluation of report generation, all available at \href{https://github.com/rajpurkarlab/CRIMSON}{https://github.com/rajpurkarlab/CRIMSON.}

\end{abstract}

\section{Introduction}

Automated radiology report generation has advanced rapidly with the emergence of large vision-language models, yet reliable evaluation remains a fundamental challenge \cite{zhang2025automated,yu2023evaluating,li2025reevalmed}. Recent radiology-specific metrics have moved beyond surface-level text similarity and instead assess factual correctness through structured error counting and finding-level comparison \cite{ostmeier2024green,zhao2024ratescore,huang2024fineradscore,zhang2025gema,wang2024llm,chaves2024towards}. These approaches represent important progress toward clinically meaningful evaluation by explicitly detecting hallucinations and omissions.

Despite these advances, current metrics largely treat detected errors as either uniformly important or binary (significant vs. not significant), and evaluate findings in relative isolation from broader clinical context. In practice, the clinical consequences of errors vary substantially. For instance, failing to report a life-threatening pneumothorax is categorically different from missing age-related aortic calcification. Moreover, the relevance and interpretation of findings depend on a patient’s age and indication. Existing evaluation frameworks do not explicitly encode clinical severity as a fine-grained spectrum; instead, they collapse findings into coarse categories, lack sufficient clinical context to determine true significance, or rely on LLM judgments without structured in-context guidelines \cite{ostmeier2024green,huang2024fineradscore,wang2024llm}. Consequently, these frameworks conflate minor, clinically inconsequential discrepancies with omissions that directly impact patient safety.

\begin{figure}[t]
    \centering
    \includegraphics[width=0.75\linewidth]{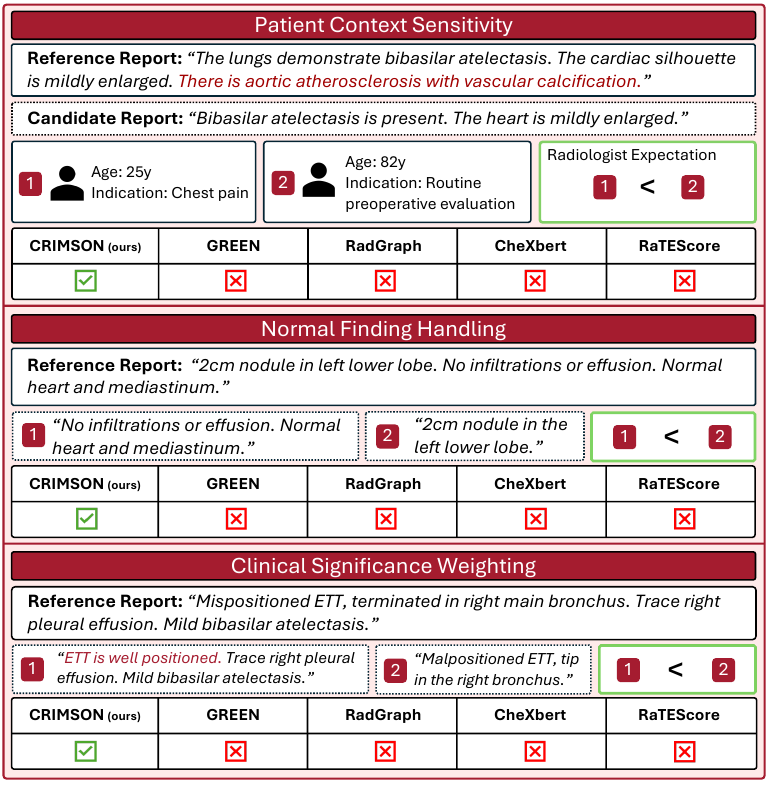}
    \caption{Representative RadJudge cases illustrating core design principles of CRIMSON. \textbf{Top:} Patient context sensitivity. The clinical impact of an omission (e.g., aortic atherosclerosis) varies by age and indication, and CRIMSON adjusts severity accordingly. \textbf{Middle:} Normal finding handling. CRIMSON does not reward mentioning normal findings, preventing score inflation. \textbf{Bottom:} Clinical significance weighting. Errors are weighted by consequence, prioritizing clinically important findings. In each case, CRIMSON aligns with radiologist expectations, whereas prior metrics fail.}
    \label{fig:figureone}
\end{figure}

To address these limitations, we introduce CRIMSON, a clinically grounded LLM-based evaluation framework designed to align automated assessment with real-world radiologic reasoning. CRIMSON evaluates reports at the level of individual findings while incorporating full clinical context, including patient age and indication. The framework models a comprehensive taxonomy of errors—false findings, missing findings, attribute-level errors (e.g., location, severity, measurement) and significance labels (urgent, actionable non-urgent, non-actionable, or expected/benign)—defined according to a rubric developed with cardiothoracic radiologists. The severity labels determine   weights within a principled scoring formulation that prioritizes clinically consequential errors over benign discrepancies and supports partial credit for partially correct findings under fine-grained attribute rules. We validate CRIMSON through alignment with radiologist-annotated clinically significant error counts in ReXVal \cite{yu2023radiology,yu2023evaluating} (PhysioNet Credentialed Health Data License 1.5.0). We also introduce and validate the metric using RadJudge, a targeted ranking test suite of clinically challenging scenarios, and RadPref, a comprehensive radiologist preference benchmark, demonstrating improved agreement with expert judgment compared to existing metrics. To facilitate reproducibility and adoption, we publicly release the metric and additionally fine-tune MedGemma \cite{sellergren2025medgemma} to generate CRIMSON predictions, enabling privacy-preserving, fully local deployment for hospitals and institutions without transmitting patient reports to external APIs.

\section{Related Work}

\textbf{Radiology Report Evaluation.} Early evaluation of radiology report generation relied on general-purpose metrics such as BLEU \cite{papineni2002bleu}, ROUGE \cite{lin2004rouge}, and METEOR \cite{banerjee2005meteor}, which measure lexical overlap and are poorly aligned with clinical correctness. Radiology-specific frameworks such as CheXbert \cite{smit2020chexbert} and RadGraph \cite{jain2021radgraph} shift evaluation toward label-based or entity-based comparisons; however, they remain constrained by predefined label spaces or extraction pipelines and do not explicitly model variation in clinical severity across findings. BERTScore \cite{zhang2019bertscore} offers embedding-based semantic similarity but does not capture clinical correctness and lacks interpretability. RadCliQ \cite{yu2023evaluating} combines multiple automated metrics to approximate radiologist judgment through a composite score. RaTEScore \cite{zhao2024ratescore} evaluates reports using entity-aware semantic similarity to better align with clinical content. Both metrics, however, lack explicit finding-level interpretability and do not incorporate clinical context to modulate error severity.

\textbf{LLM-Based Evaluation Frameworks.} More recently, evaluation methods have leveraged large language models (LLMs) to assess factual correctness through error categorization and counting \cite{ostmeier2024green,wang2024llm,huang2024fineradscore,li2025s,bannur2024maira,zhang2025gema,jiang2025clear,dua2025clinically}. Compared to label- or graph-based approaches, LLM-based evaluators offer greater flexibility in identifying nuanced discrepancies and can produce structured rationales alongside error counts, improving interpretability and transparency. GREEN \cite{ostmeier2024green} and LLM-RadJudge \cite{wang2024llm} both leverage LLMs to identify and classify errors in candidate reports, producing interpretable evaluations validated against radiologist judgments. However, these and most other LLM-based metrics either treat all errors uniformly or rely on a binary significant versus insignificant distinction, without modeling clinical severity along a fine-grained spectrum or incorporating full patient-level context \cite{ostmeier2024green,wang2024llm,li2025s,zhang2025gema}. FineRadScore \cite{huang2024fineradscore} introduces error weighting within a line-level correction framework; however, its severity assessment remains implicitly determined by the LLM without explicit clinician-defined guidelines or comprehensive patient-level context. Other LLM-based metrics take alternative approaches: CLEAR \cite{jiang2025clear} transforms reports into structured condition--attribute tables for attribute-level fidelity assessment, while ICARE \cite{dua2025clinically} employs a dual-agent question-answering framework for clinically grounded evaluation. Zhu et al.  \cite{zhu2024leveraging} demonstrate that incorporating professional radiologists' expertise into LLM-based evaluation pipelines can substantially improve alignment with clinical judgment.

\section{CRIMSON: Clinically-Grounded Report Evaluation}

CRIMSON evaluates a candidate radiology report against a reference report by identifying and categorizing discrepancies at the finding level, weighing each error by its clinical significance, and computing a normalized score that reflects clinical preferences. GPT-5.2 \cite{singh2025openai} is used as the backbone throughout this pipeline. The framework operates in three stages: (1) finding extraction and clinical significance assignment, (2) error detection and classification, and (3) severity-aware score computation.

\subsection{Finding Extraction and Clinical Significance Assignment}

Given a reference report $R_{\text{ref}}$ and a candidate report $R_{\text{cand}}$, CRIMSON first extracts all abnormal findings from each report. Normal findings are excluded from evaluation because including them can introduce spurious variability due to stylistic differences across radiologists \cite{baharoon2025radgame,baharoon2025exploring}. Each extracted finding $f$ is assigned a clinical significance weight $w(f)$ based on standard radiological practice and a structured severity framework adapted from \cite{tian2023refisco} with input from attending cardiothoracic radiologists. The clinical significance weight $w(f)$ is defined according to the following:

\begin{equation}
    w(f) = \begin{cases} 1.0 & \text{if urgent} \\ 0.5 & \text{if actionable, not urgent} \\ 0.25 & \text{if not actionable, not urgent} \\ 0.0 & \text{if expected/benign} \end{cases}
\end{equation}

Findings classified as \textit{urgent} correspond to abnormalities requiring immediate intervention or indicating life-threatening conditions, such as a tension pneumothorax. \textit{Actionable non-urgent} findings are those that alter patient management but are not immediately critical, including nodules, moderate pleural effusions, or consolidations. \textit{Not actionable} are findings that represent minimal clinical impact but remain worth documenting, such as a cervical rib or appropriately positioned support devices. \textit{Expected benign} findings include expected or age-appropriate changes with no impact on care, such as degenerative spine changes or a tortuous aorta. We separate non-actionable from expected benign findings because reporting of expected benign findings varies substantially by radiologist style; penalizing based on them introduces unnecessary randomness. Clinical significance assignment incorporates patient context when available, including age and indication. For example, aortic calcification in a 75-year-old patient is classified as expected benign, whereas the same finding in a 25-year-old patient may be considered actionable non-urgent due to atypical early onset.

\subsection{Error Taxonomy and Classification}

CRIMSON characterizes discrepancies through three primary error categories: false findings, missing findings, and attribute errors. False findings are abnormal findings present in $R_{\text{cand}}$ but absent from $R_{\text{ref}}$, representing a hallucination. Missing findings are abnormal findings present in $R_{\text{ref}}$ but absent from $R_{\text{cand}}$, representing diagnostically meaningful omissions. 

Findings that appear in both reports are considered “matched.” For such findings, CRIMSON evaluates attribute-level correctness across eight dimensions: (1) anatomical location or laterality, (2) severity or extent, (3) morphological descriptors, (4) quantitative measurements, (5) certainty level, (6) diagnostic underinterpretation, (7) overinterpretation, and (8) temporal or comparison descriptors.

Each attribute error $e$ is assigned a severity-based weight $w_{\text{attr}}(e)$, where $w_{\text{attr}}(e) = 0.5$ if the error is labeled as significant, and $w_{\text{attr}}(e) = 0.0$ if it is labeled as negligible. Significant attribute errors are those that could alter treatment decisions or patient management, whereas negligible errors correspond to clinically inconsequential differences. For example, incorrect lung laterality is considered significant, whereas positional differences within the same lobe, such as `apical' vs `lateral' are considered negligible. For pulmonary nodules smaller than 6 mm, measurement discrepancies exceeding 2 mm are considered significant; for nodules 6 mm or larger, discrepancies exceeding 4 mm are considered significant, reflecting established practice \cite{macmahon2017guidelines}. Changes in severity descriptors that affect urgency, such as “small” versus “large,” are significant, whereas as “small” versus “tiny” are negligible. A single matched finding may contain multiple attribute errors, each evaluated independently.

\subsection{Severity-Aware Scoring}

The framework produces a score in the range of $(-1,1]$ that can be easily interpreted in clinical workflows. The scale is grounded at $0$, which corresponds to a normal candidate report. This reflects a practical assumption that a radiologist begins from a normal template and modifies the report by adding abnormal findings. A score greater than zero indicates that the candidate report contains more correct findings than errors after severity weighting. A score equal to zero indicates that the report is no more informative than submitting a normal template, except when the reference report is also normal, in which case a correct normal report receives a score of $1$. A score less than zero indicates that the report contains more errors than correct findings, implying that a radiologist would likely spend more effort correcting it than editing a normal template. The upper bound of $1$ represents a perfect report with no missed findings, no false positives, and no significant attribute errors. Negative scores approach $-1$ asymptotically because errors are theoretically unbounded: a candidate can always become worse by introducing additional false findings.

For each matched finding $m_i$, let its clinical significance weight be 
$w_i = w(m_i)$. Attribute-level penalties are aggregated as 
$E_{\text{attr}, i} = \sum_j w_{\text{attr}}(e_{j,i})$, where $e_{j,i}$ referes to attribute error $j$ for finding $i$, and total credit, $C$, across matched findings is 
$C = \sum_{i \in \text{matched}} w_i \cdot \frac{w_i}{w_i + E_{\text{attr}, i}}$.
Let $W_{\text{ref}} = \sum_{f \in R_{\text{ref}}} w(f)$ denote the total weighted clinical significance of the reference report, and let $E_{\text{false}} = \sum_{f \in \text{false}} w(f)$ denote the weighted sum of false positive findings. The raw score is defined as:

\begin{equation}
S =
\begin{cases}
\dfrac{C - E_{\text{false}}}{W_{\text{ref}}} 
& \text{if } W_{\text{ref}} > 0, \\[8pt]

- E_{\text{false}} 
& \text{if } W_{\text{ref}} = 0 \text{ and } E_{\text{false}} > 0. \\

1 
& \text{if } W_{\text{ref}} = 0 \text{ and } E_{\text{false}} = 0, \\[8pt]

\end{cases}
\end{equation}

To bound the negative range while preserving relative ordering, let $A = E_{\text{false}} - C$, which represents the excess weighted errors relative to correct findings. The final score is:

\begin{equation}
\text{CRIMSON} =
\begin{cases}
S & \text{if } S \ge 0 \\
-\dfrac{A}{1 + A} & \text{if } S < 0 
\end{cases}
\end{equation}

\section{Results} 

We perform three complementary forms of validation: correlation with radiologist-annotated clinically significant error counts (Section~\ref{sec:correlation_errors}), a radiologist-guided pass–fail clinical judgment test on RadJudge (Section~\ref{sec:radjudge}), and large-scale radiologist preference alignment on RadPref (Section~\ref{sec:radpref}). All metrics except CRIMSON were computed using RadEval \cite{xu2025radeval}.

\subsection{Correlation with Radiologist-Annotated Significant Errors}\label{sec:correlation_errors}

We evaluated CRIMSON on 50 cases from ReXVal \cite{yu2023radiology} annotated by six board-certified radiologists. We computed Kendall’s $\tau$ and Pearson $r$ correlations between automatic metric scores and radiologist-derived clinically significant error counts. As shown in Table~\ref{tab:correlation_multi_reference}, CRIMSON demonstrates strong alignment with these expert annotations. Furthermore, error counts ($E$) exhibited even stronger alignment, and severity-weighted errors (Weighted $E$) achieved the highest correlations overall, demonstrating that explicitly modeling clinical consequence further improves agreement with expert judgment.

\begin{table*}
\centering
\caption{
Kendall $\tau$ and Pearson $r$ correlations (95\% CI) between automatic metrics and radiologist-derived clinically significant error counts ($n=50$). Columns refer to different candidate reports on ReXVal, each of which was chosen to optimize a specific metric \cite{yu2023radiology}.
GREEN $E$ and CRIMSON $E$ denote the total (unweighted) error count, while CRIMSON Weighted $E$ applies clinical severity–based weighting to errors. All other metrics were calculated using RadEval \cite{xu2025radeval}. $^{*}$CRIMSON results are averaged across 5 runs due to non-deterministic API outputs.
}
\label{tab:correlation_multi_reference}

\fontsize{8}{10}\selectfont
\setlength{\tabcolsep}{4pt}

\begin{tabular}{lcccc}
\toprule
& \multicolumn{4}{c}{\textbf{Correlation with Significant Error Counts}} \\
\cmidrule(lr){2-5}
\textbf{Metric}
& \multicolumn{2}{c}{\textbf{CheXbert}}
& \multicolumn{2}{c}{\textbf{BERTScore}}\\
\cmidrule(lr){2-3} \cmidrule(lr){4-5}
& Kendall $\tau$ & Pearson $r$ & Kendall $\tau$ & Pearson $r$  \\
\midrule

RadGraph \cite{jain2021radgraph}
& $0.41\,\ci{0.19}{0.61}$ & $0.59\,\ci{0.41}{0.75}$
& $0.54\,\ci{0.36}{0.68}$ & $0.65\,\ci{0.51}{0.78}$ \\

BLEU \cite{papineni2002bleu}
& $0.49\,\ci{0.30}{0.65}$ & $0.60\,\ci{0.47}{0.72}$
& $0.36\,\ci{0.16}{0.54}$ & $0.48\,\ci{0.32}{0.63}$ \\

BERTScore \cite{zhang2019bertscore}
& $0.52\,\ci{0.35}{0.67}$ & $0.65\,\ci{0.52}{0.78}$
& $0.49\,\ci{0.30}{0.66}$ & $0.60\,\ci{0.44}{0.74}$ \\

GREEN \cite{ostmeier2024green}
& $0.62\,\ci{0.46}{0.75}$ & $0.75\,\ci{0.64}{0.86}$
& $0.67\,\ci{0.54}{0.78}$ & $0.70\,\ci{0.59}{0.80}$ \\

ROUGE-L \cite{lin2004rouge}
& $0.58\,\ci{0.44}{0.71}$ & $0.71\,\ci{0.60}{0.81}$
& $0.54\,\ci{0.37}{0.70}$ & $0.62\,\ci{0.46}{0.75}$ \\

CheXbert \cite{smit2020chexbert}
& $0.46\,\ci{0.26}{0.63}$ & $0.45\,\ci{0.18}{0.70}$
& $0.30\,\ci{0.08}{0.51}$ & $0.34\,\ci{0.09}{0.60}$ \\

RaTEScore \cite{zhao2024ratescore}
& $0.39\,\ci{0.17}{0.57}$ & $0.52\,\ci{0.31}{0.69}$
& $0.49\,\ci{0.32}{0.65}$ & $0.56\,\ci{0.37}{0.73}$ \\

RadCliQ-v1 \cite{yu2023evaluating}
& $0.34\,\ci{0.21}{0.46}$ & $0.34\,\ci{0.19}{0.52}$
& $0.35\,\ci{0.21}{0.48}$ & $0.35\,\ci{0.22}{0.53}$ \\

CRIMSON$^{*}$
& $0.68\,\ci{0.54}{0.79}$ & $0.84\,\ci{0.76}{0.90}$
& $0.71\,\ci{0.60}{0.80}$ & $0.82\,\ci{0.74}{0.89}$ \\

\midrule

GREEN $E$
& $0.71\,\ci{0.59}{0.81}$ & $0.75\,\ci{0.65}{0.86}$
& $0.75\,\ci{0.63}{0.85}$ & $0.85\,\ci{0.75}{0.92}$ \\

CRIMSON $E$
& $0.73\,\ci{0.61}{0.83}$ & $0.88\,\ci{0.79}{0.94}$
& $0.72\,\ci{0.62}{0.80}$ & $0.86\,\ci{0.77}{0.93}$ \\

CRIMSON Weighted $E$
& $\mathbf{0.78\,\ci{0.67}{0.86}}$ & $\mathbf{0.90\,\ci{0.85}{0.95}}$
& $\mathbf{0.80\,\ci{0.71}{0.87}}$ & $\mathbf{0.91\,\ci{0.88}{0.95}}$ \\

\bottomrule
\end{tabular}

\begin{tabular}{lcccc}
& \multicolumn{2}{c}{\textbf{RadGraph}}
& \multicolumn{2}{c}{\textbf{BLEU}}\\
\cmidrule(lr){2-3} \cmidrule(lr){4-5}
& Kendall $\tau$ & Pearson $r$ & Kendall $\tau$ & Pearson $r$  \\
\midrule

RadGraph
& $0.59\,\ci{0.46}{0.71}$ & $0.60\,\ci{0.50}{0.72}$
& $0.64\,\ci{0.50}{0.75}$ & $0.75\,\ci{0.65}{0.84}$ \\

BLEU
& $0.13\,\ci{0.09}{0.34}$ & $0.23\,\ci{0.02}{0.39}$
& $0.52\,\ci{0.34}{0.68}$ & $0.67\,\ci{0.53}{0.79}$ \\

BERTScore
& $0.46\,\ci{0.29}{0.61}$ & $0.54\,\ci{0.39}{0.68}$
& $0.58\,\ci{0.41}{0.72}$ & $0.72\,\ci{0.60}{0.83}$ \\

GREEN
& $0.62\,\ci{0.48}{0.74}$ & $0.65\,\ci{0.53}{0.77}$
& $0.70\,\ci{0.55}{0.83}$ & $0.79\,\ci{0.68}{0.89}$ \\

ROUGE-L
& $0.54\,\ci{0.38}{0.67}$ & $0.60\,\ci{0.49}{0.70}$
& $0.67\,\ci{0.52}{0.79}$ & $0.80\,\ci{0.70}{0.87}$ \\

CheXbert
& $0.29\,\ci{0.08}{0.48}$ & $0.33\,\ci{0.08}{0.55}$
& $0.18\,\ci{0.05}{0.40}$ & $0.23\,\ci{0.04}{0.47}$ \\

RaTEScore
& $0.57\,\ci{0.42}{0.70}$ & $0.62\,\ci{0.45}{0.78}$
& $0.54\,\ci{0.39}{0.68}$ & $0.67\,\ci{0.54}{0.79}$ \\

RadCliQ-v1
& $0.12\,\ci{0.05}{0.28}$ & $0.06\,\ci{0.11}{0.28}$
& $0.28\,\ci{0.11}{0.43}$ & $0.16\,\ci{0.01}{0.53}$ \\

CRIMSON$^{*}$
& $0.61\,\ci{0.45}{0.75}$ & $0.71\,\ci{0.53}{0.85}$
& $0.67\,\ci{0.54}{0.79}$ & $0.81\,\ci{0.71}{0.89}$ \\

\midrule

GREEN $E$
& $0.71\,\ci{0.59}{0.81}$ & $0.75\,\ci{0.65}{0.86}$
& $\mathbf{0.80\,\ci{0.71}{0.88}}$ & $\mathbf{0.88\,\ci{0.82}{0.93}}$ \\

CRIMSON $E$
& $0.73\,\ci{0.61}{0.83}$ & $0.86\,\ci{0.78}{0.92}$
& $0.74\,\ci{0.61}{0.84}$ & $0.87\,\ci{0.78}{0.93}$ \\

CRIMSON Weighted $E$
& $\mathbf{0.77\,\ci{0.67}{0.85}}$ & $\mathbf{0.86\,\ci{0.80}{0.93}}$
& $0.78\,\ci{0.69}{0.86}$ & $\mathbf{0.88\,\ci{0.82}{0.93}}$ \\

\bottomrule
\end{tabular}
\end{table*}

\subsection{Radiologist-Guided Clinical Judgment Test}\label{sec:radjudge}

\begin{figure}[t]
    \centering
    \includegraphics[width=0.8\linewidth]{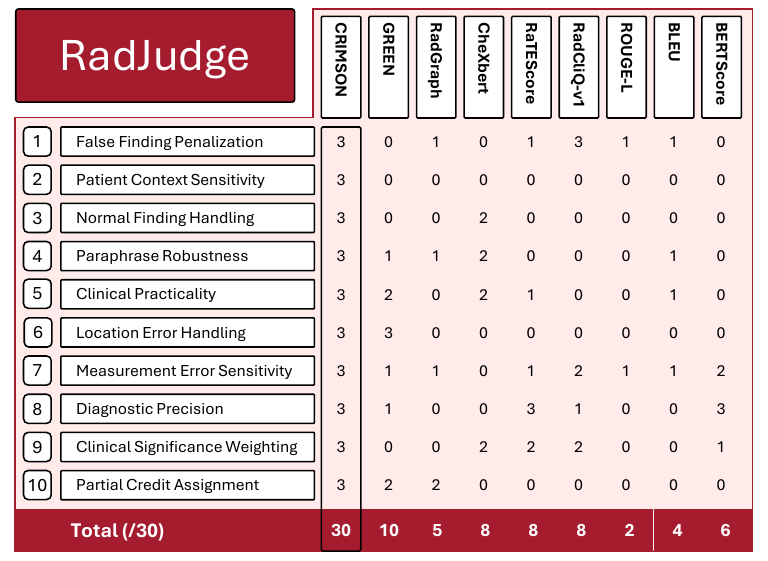}
    \caption{RadJudge results. For each case, metrics are evaluated based on whether their relative ranking of multiple candidate reports agrees with the expected ordering determined with agreement across three attending cardiothoracic radiologists. Each category contains three cases; entries are cases passed (out of 3), with totals out of 30.}
\label{fig:figuretwo}
\end{figure}

\begin{figure}[t]
    \centering
    \includegraphics[width=0.9\linewidth]{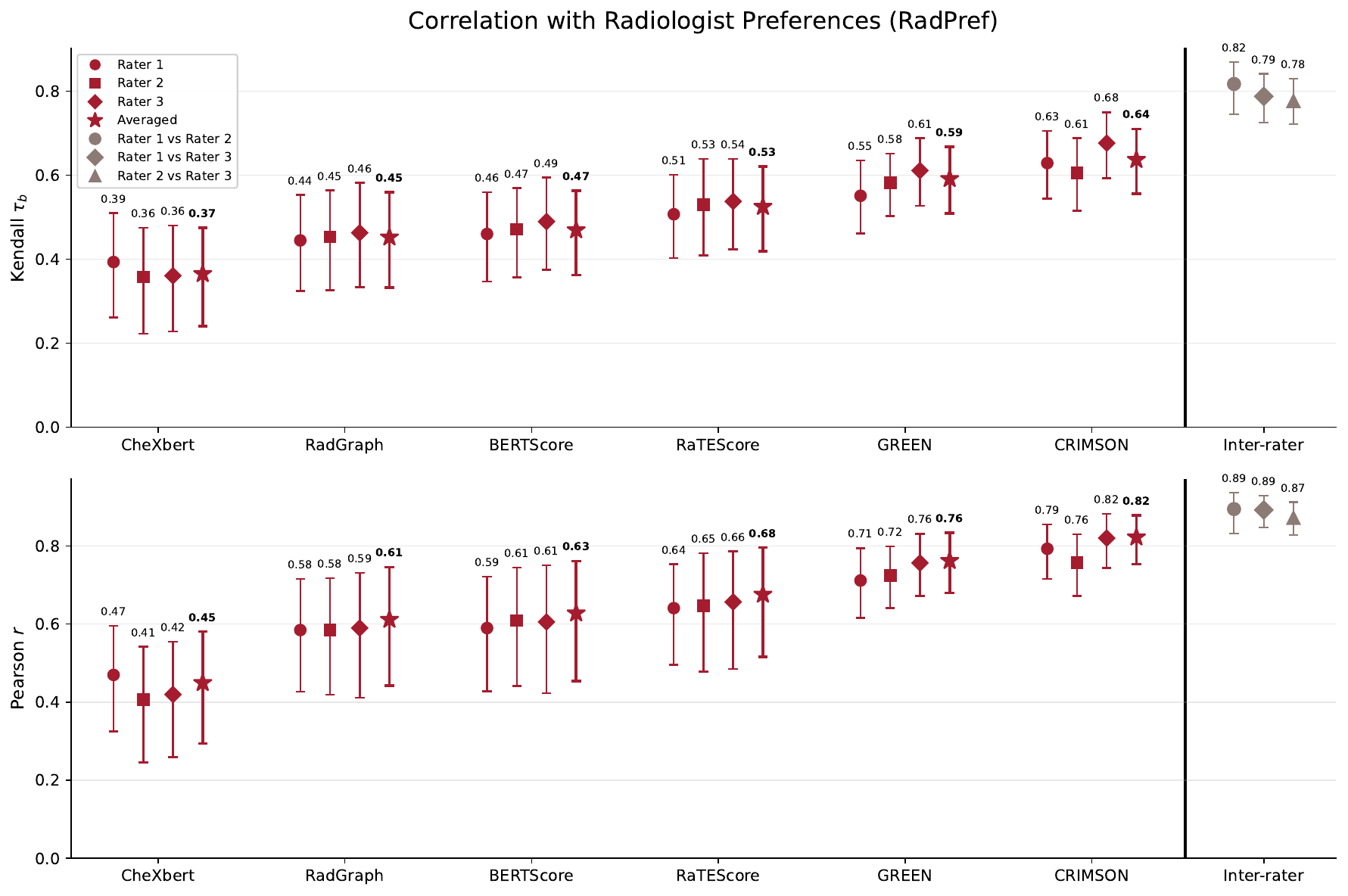}
    \caption{Radiologist Preference Alignment (RadPref). Correlation between metric score and radiologist rating differences across 100 pairwise cases. Each point corresponds to a case comparing two candidate reports for the same reference report.}
\label{fig:figurethree}
\end{figure}

We also developed RadJudge, a targeted pass–fail test suite reflecting real-world radiologist intuition. RadJudge comprises 30 curated cases across 10 clinically nuanced categories in which multiple candidate reports are compared and independently reviewed by three cardiothoracic radiologists, with agreement required to establish the reference preference. A metric passes if it ranks reports in accordance with expert judgment or assigns equivalent scores when radiologists deem them clinically indistinguishable, defined as differences within a threshold of 0.01. Three representative cases are shown in Figure~\ref{fig:figureone}. The suite probes challenging scenarios, including urgent omissions versus benign hallucinations, context-dependent findings, diagnostic over- and under-interpretation, and situations reflecting the clinical reality of imperfect reference reports that omit localization or age-expected benign findings.

As shown in Figure \ref{fig:figuretwo}, CRIMSON is the only metric that correctly solves all 30 out of 30 cases, consistently ranking candidate reports in accordance with expert radiologist judgment. In contrast, all prior metrics perform substantially worse, correctly resolving fewer than 35\% of cases, highlighting their limited ability to capture nuanced clinical judgment.

\subsection{Radiologist Preference Alignment}\label{sec:radpref}

Correlation with error counts may not fully capture radiologist preference, as experts don't weigh different types of errors equally in overall judgment. To directly assess preference alignment, we introduce RadPref, a benchmark of 100 cases, each containing a reference report from ReXGradient-160K \cite{zhang2025rexgradient} and two candidate reports randomly generated using diverse regimes inspired by \cite{ostmeier2024green}: report generation with MedGemma \cite{sellergren2025medgemma}, randomly sampled reports, BERT similarity–matched reports, and LLM-based editing, addition, or removal of findings. Each candidate was rated on a 1–5 scale by three cardiothoracic radiologists based on overall clinical quality and correctness relative to the reference report. The scale was defined as: 1 = completely wrong or clinically dangerous; 2 = major errors, with most key findings missing or false; 3 = partially correct, with some significant errors; 4 = mostly accurate, with only minor or negligible errors; and 5 = clinically equivalent to the ground truth. Scores were computed separately for both candidates.

\begin{figure}
    \centering
    \includegraphics[width=1\linewidth]{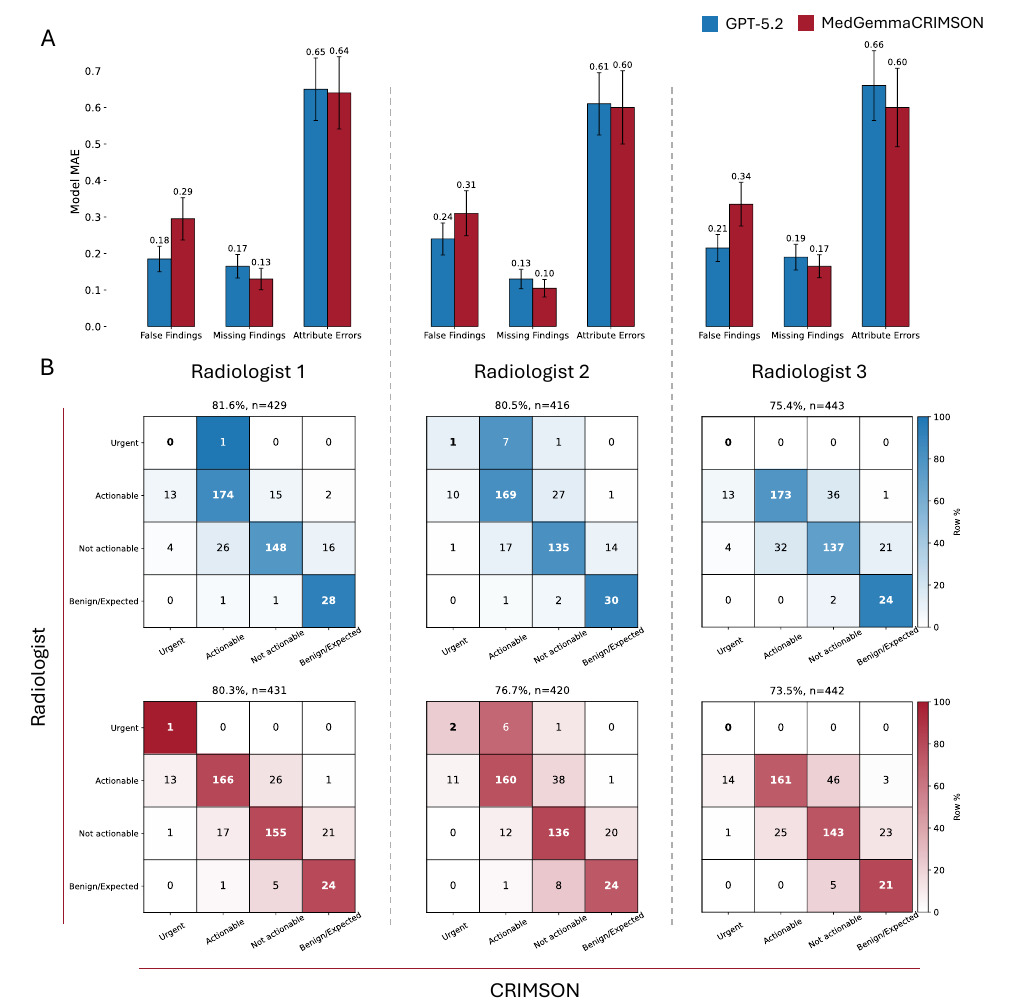}
    \caption{MedGemmaCRIMSON vs GPT-5.2. 
    A) Mean absolute error across false findings, missing findings, and attribute errors per radiologist. 
    B) Severity categorization confusion matrices between three radiologists and CRIMSON, computed only on matched errors (i.e., findings for which both the radiologist and CRIMSON identified an error in the same category). Titles show the percentage of cases for which the radiologist and CRIMSON agree on error category. Color intensity represents the within-row percentage.}
\label{fig:figurefour}
\end{figure}

Figure~\ref{fig:figurethree} shows that CRIMSON demonstrates the strongest alignment with radiologist pairwise preferences among all evaluated metrics. Across Kendall’s $\tau_b$ and Pearson $r$, CRIMSON consistently outperforms prior approaches and approaches inter-rater radiologist agreement. These findings indicate that CRIMSON more faithfully reflects expert clinical preference in relative report quality.

\subsection{MedGemma Fine-tuning and Analysis}\label{sec:medgemma}

To enable privacy-preserving, fully local deployment of CRIMSON, we fine-tuned MedGemma \cite{sellergren2025medgemma} on GPT-5–generated CRIMSON annotations using the full ReXGradient-160K \cite{zhang2025rexgradient} training set of 140{,}000 reports pairs for 10 epochs using LoRA \cite{hu2022lora}. For each pair, the candidate report was generated using the same regimes described in Section~\ref{sec:radpref} except without using any image-based generation. GPT-5 generated structured finding-level error labels and severity assignments, which served as supervision to train MedGemma to replicate CRIMSON-style outputs. Additional training details are provided on the model’s Hugging Face page.

We compare the fine-tuned MedGemma (MedGemmaCRIMSON) against GPT-5.2 on RadPref preference alignment and severity categorization (Figure~\ref{fig:figurefour}). Notably, RadPref provides not only pairwise preference ratings but also structured error categorization and clinical severity annotations across all three cardiothoracic radiologists, enabling direct evaluation of both preference alignment and clinically significant error modeling. MedGemmaCRIMSON achieves comparable mean absolute error to GPT-5.2 across false findings, missing findings, and attribute errors, with similar behavior particularly on attribute-level discrepancies. For clinical significance labeling, MedGemmaCRIMSON closely mirrors GPT-5.2 in reproducing radiologist-assigned categories across all three radiologists, achieving agreement rates that are slightly lower but within a narrow margin of GPT-5.2 (Radiologist 1: 80.3\% vs 81.6\%; Radiologist 2: 76.7\% vs 80.5\%; Radiologist 3: 73.5\% vs 75.4\%), with most disagreements occurring between adjacent severity levels rather than extreme misclassifications.

\section{Discussion}

We introduce CRIMSON, a clinically grounded and severity-aware framework for fine-grained radiology report evaluation that explicitly models patient context, diagnostic consequence, and structured attribute-level errors. By incorporating clinician-defined clinical significance weights and score normalization, CRIMSON aligns automated evaluation more closely with real-world radiologist reasoning than prior approaches. While CRIMSON leverages GPT-5.2 for structured evaluation, we additionally demonstrate that a fine-tuned open-weight model (MedGemmaCRIMSON) can closely approximate its behavior, enabling privacy-preserving, fully local deployment.

A core motivation of CRIMSON is the principle that generated reports should be evaluated according to how they would function under radiologist oversight, rather than solely through aggregate accuracy or raw error counts. Instead of treating all discrepancies equally or as binary (significant vs.\ not significant), CRIMSON explicitly models whether an error would be clinically consequential or potentially dangerous. Missing a life-threatening abnormality should dominate the evaluation, whereas minor descriptor differences should not.

This principle also motivates CRIMSON’s partial-credit design for attribute errors. When a model correctly identifies a clinically important finding but misstates a secondary attribute (e.g., mild severity mismatch or imprecise localization), it may still provide value by directing the radiologist’s attention to the relevant abnormality. CRIMSON therefore rewards correct detection while penalizing clinically meaningful attribute mistakes in a severity-aware manner, reflecting that some errors increase downstream review effort without necessarily creating the same patient-safety risk as a complete omission or major hallucination.

Across three complementary validation settings: (1) correlation with radiologist-annotated clinically significant error counts, (2) the RadJudge clinical judgment suite, and (3) the RadPref radiologist preference benchmark, CRIMSON consistently demonstrates stronger agreement with expert judgment than existing metrics. Notably, severity-weighted modeling further improves alignment, highlighting the importance of distinguishing clinically consequential errors from benign ones.

A limitation of this work is that much of CRIMSON’s prompting framework, severity rubric, and structured evaluation guidelines were developed specifically for chest X-ray reports. The clinical significance taxonomy, attribute rules, and measurement thresholds were designed in collaboration with cardiothoracic radiologists and tailored to common CXR findings and reporting conventions. Although the underlying evaluation framework is modality-agnostic in principle, applying CRIMSON to other imaging domains will require adaptation of prompts, finding ontologies, and severity criteria to align with modality-specific clinical standards. Future work will extend CRIMSON beyond chest X-ray to additional imaging modalities where anatomical detail, multimodal context, and diagnostic complexity are considerably greater.

\section*{Acknowledgments}
This research was supported in part by Harvard Medical School Dean’s Innovation Award for Accelerating Foundation Model Research.

\bibliographystyle{splncs04}
\bibliography{references}
\end{document}